\documentclass[runningheads]{llncs}
\usepackage{graphicx}
\usepackage{comment}
\usepackage{amsmath,amssymb} 
\usepackage{color}
\usepackage{epsfig}
\usepackage{amsmath}
\usepackage{amssymb}

\usepackage{xfrac}
\usepackage{bm}
\usepackage{caption}
\usepackage{subcaption}
\usepackage{float}
\usepackage{color,soul}
\usepackage{xcolor}
\usepackage{mathtools}    
\usepackage{multirow, makecell}
\usepackage[utf8]{inputenc}

\usepackage{threeparttable}
\usepackage{booktabs}

\usepackage[pagebackref=true,breaklinks=true,letterpaper=true,colorlinks,bookmarks=false]{hyperref}

\newcommand{\cL}{\mathcal{L}}

\definecolor{nicegreen}{HTML}{99C000}
\definecolor{nicegreen2}{HTML}{81d41a}
\definecolor{niceyellow}{HTML}{FDCA00}
\definecolor{niceorange}{HTML}{f5a300}
\definecolor{niceblue}{HTML}{0083cc}
\definecolor{nicepurple}{HTML}{a60084}

\begin{document}
\pagestyle{headings}
\mainmatter

\title{NODIS: Neural Ordinary Differential \\ Scene Understanding} 

\titlerunning{NODIS}
%
\author{Yuren Cong\inst{1} \and Hanno Ackermann\inst{1} \and Wentong Liao\inst{1} \and \\
Michael Ying Yang\inst{2} \and Bodo Rosenhahn\inst{1}}
\authorrunning{Cong et al.}
%
\institute{Institute of Information Processing, Leibniz University Hannover, Germany \\
\email{\{cong, ackermann, liao, rosenhahn\}@tnt.uni-hanover.de}
\and
Scene Understanding Group, University of Twente, The Netherlands
\email{michael.yang@utwente.nl}}
\maketitle


\begin{abstract}
    Semantic image understanding is a challenging topic in computer vision. It requires to detect all objects in an image, but also to identify all the relations between them. Detected objects, their labels and the discovered relations can be used to construct a scene graph which provides an abstract semantic interpretation of an image. In previous works, relations were identified by solving an assignment problem formulated as (Mixed-)Integer Linear Programs. 
    In this work, we interpret that formulation as Ordinary Differential Equation (ODE).  The proposed architecture performs scene graph inference by solving a neural variant of an ODE by end-to-end learning. The connection between (Mixed-)Integer Linear Program and ODEs in combination with the end-to-end training amounts to learning how to solve assignment problems with image-specific objective functions. Intuitive, visual explanations are provided for the role of the single free variable of the ODE modules which are associated with time in many natural processes. 
    The proposed model achieves results equal to or above state-of-the-art on all three benchmark tasks: scene graph generation (SGGEN), classification (SGCLS) and visual relationship detection (PREDCLS) on Visual Genome benchmark. The strong results on scene graph classification support the claim that assignment problems can indeed be solved by neural ODEs.
\end{abstract}


\section{Introduction}

This paper investigates the problem of semantic image understanding. Given an image, the objective is to detect objects within, label them and infer the relations which might exist between objects. These data provide rich semantic information about the image content. 
So called scene graphs contain all these information and constitute abstract representations of images~\cite{johnson2015image}. Nodes of a scene graph represent objects detected in an image, while edges represent relationships between objects. 
Applications range from image retrieval~\cite{johnson2015image} to high-level vision tasks such as visual question answering~\cite{teney2017graph}, image captioning~\cite{yao2018exploring,yang2019auto} and visual reasoning~\cite{shi2019explainable}. The community has been very active in the past years to investigate these problems. 
Results on public benchmarks such as the Visual Genome database~\cite{krishna2017visual} have improved drastically within the past few years~\cite{xu2017scene,zellers2018neural,chen2019counterfactual}.


A naive approach to infer scene graphs is to use a standard object detector, classify the objects, and use a separate network to label the relationships 
\begin{figure}[t!]
\centering
\includegraphics[width=0.975\linewidth]{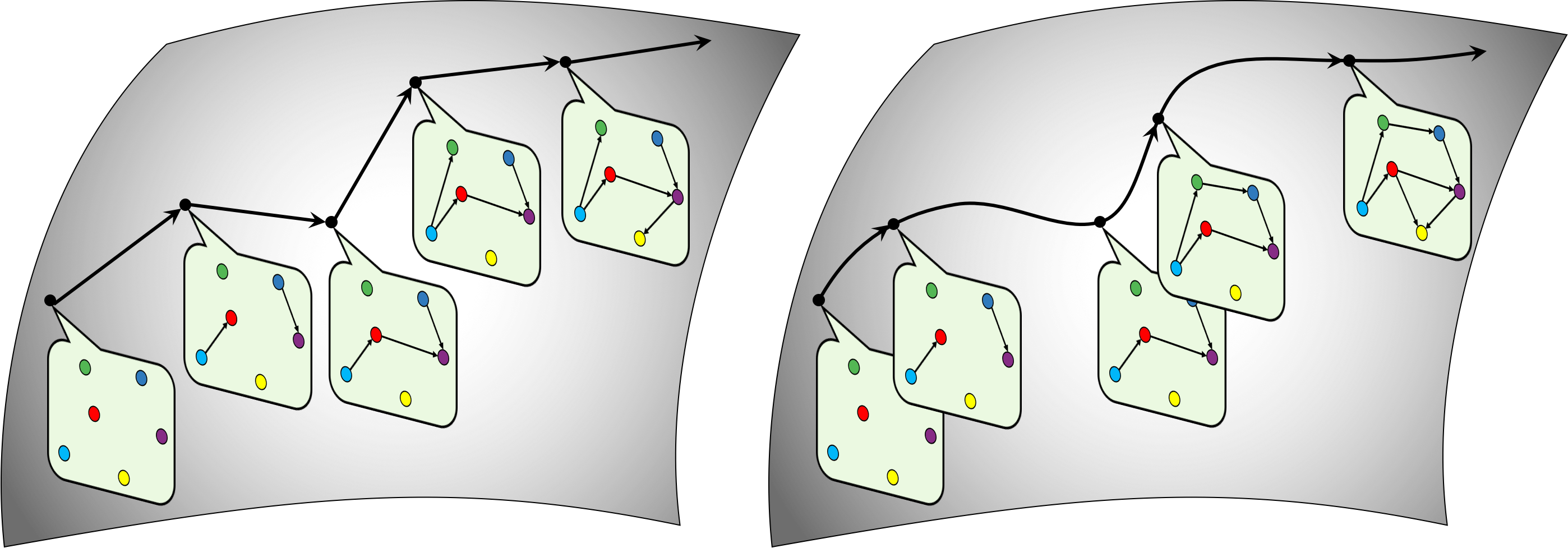}
\caption{Visualization of the main contributions made in this work. \textit{Left}: state-of-the-art works rely on multiple embedding layers which amounts to an evenly-spaced, discrete sampling. \textit{Right}: the proposed module relies on ordinary differential equations (ODE), thus it learns a continuous approximation of the trajectory onto the embedding manifolds.}
\label{fig:teaser}
\end{figure}   
between the objects.  
This, however, ignores the conditional information shared between  the objects. 
It has therefore become standard practice to also use relationship information for object classification~\cite{hu2018relation,liu2018structure,li2017scene}.

Scene graph estimation relies on two classification problems, both for objects and for relationships. Classification, or more generally, assignment problems
can be solved by (Mixed) Integer Linear Programs, (M)ILP, for instance  in~\cite{Silberman2012:SupportInference}. Powerful solvers are able to even infer globally optimal solutions for many types of problems. 
A drawback of (M)ILPs is that the optimization objective and all constraints must be defined in advance. Objective functions usually include multiple terms and weights per term. Furthermore,  backpropagation through an (M)ILP solver is not possible, thus such pipelines cannot adapt to data. 

For many other problems in which 
adaption to data is less important, (M)ILPs have been successfully employed. 
In~\cite{Fuegenschuh2006:CombinatorialModels}, network flows and transport problem are modeled by Partial Differential Equations (PDEs). 
The arising systems of PDEs are transformed to systems of Ordinary Differential Equations (ODEs) by finite differences. Adding constraints such as minimal or maximal capacities, integral constraints and an objective, for instance energy minimization, and further applying a piece-wise linear approximation, a solution of the ODE can be obtained by means of an (M)ILP. 
That implies that for properly constructed (M)ILPs there are systems of  ODEs 
which include the solution of the (M)ILPs, since dropping the optimization objective -- thereby switching from an optimization problem to solving an equation system -- only increases the number of solutions. While backpropagating through an (M)ILP is generally not possible, backpropagation through an ODE is possible using a recent seminal work \emph{Neural Ordinary Differential Equations}~\cite{Chen2018:NODE}.

Using this link, we propose to solve the labeling problem by means of neural ordinary differential equations. This module can be interpreted such that it can learn to perform the same task as an (M)ILP. Since it allows end-to-end training, it adapts to the data, thus it learns a near-optimal objective per image.
A further question is which role the time-variable of the ODE plays in our architecture. We will provide visual explanations of that variable, namely that it controls how many objects are classified and how many relationships are classified (cf. Figs.~\ref{fig:teaser} and \ref{fig:ode_vary}).  
In our experimental evaluation, we demonstrate results that are equal to state-of-the-art or above in all three benchmark problems: scene graph generation (SGGEN), scene graph classification (SGCLS) and visual relationship prediction (PREDCLS). 

The \textbf{contributions} made in this work can be summarized as follows:
(1) We propose to use ODE modules for scene graph generation. 
(2) It can be interpreted that it learns the optimal assignment function per image. 
(3) Intuitive, visual explanations are provided for the role of the single free variable of the ODE modules which are associated with time in many natural processes.
(4) The proposed method achieves state-of-the-art results. Our code is published on GitHub \footnote{\emph{https://github.com/yrcong/NODIS}}.



\section{Related Work} 
\label{Sec:Related}

\noindent \textbf{Context for Visual Reasoning:} Context has been used in semantic image understanding 
\cite{divvala2009empirical,ladicky2010graph,yao2010modeling,hu2018relation,liu2018structure,Liao_2019_CVPR_Workshops}.
For scene graph generation, context information has been recently proposed and is still being investigated. 
Message passing has been used to include object context in images in several works, for instance by graphical models \cite{li2018factorizable,yang2018graph}, by recurrent neural networks (RNN) \cite{zellers2018neural,wang2019exploring}, or by an iterative refinement process \cite{xu2017scene,krishna2018referring}. 
Context from language priors \cite{mikolov2013efficient} has been proved to be helpful for visual relationships detection and scene graph generation \cite{lu2016visual,yu2017visual,li2017scene}.

\noindent \textbf{Scene Graph Generation:} Scene graphs are proposed in \cite{johnson2015image} for the task of image retrieval and also potential for many applications \cite{kluger2019temporally,reinders2019learning,kluger2020consac}. 
They consist of not only detected objects but also the object classes and the relationships between the detected objects. 
The estimation of scene graphs from images is attracting more and more attention in computer vision \cite{li2017scene,dai2017detecting,liang2017deep,zhuang2017towards,li2018factorizable,zellers2018neural,wang2019exploring,hu2019exploiting}. 
Several of these methods use message passing to capture the context of the two related objects \cite{li2017vip,zellers2018neural,wang2019exploring}, or of the objects and their relationships \cite{xu2017scene,li2017scene,li2018factorizable,yang2018graph}.
The general pipeline for message passing is to train some shared matrices to transform the source features into a semantic domain and then to assemble them to update the target features. 
In some works, an attention mechanism is implemented to weight the propagated information to achieve further improvement \cite{li2018factorizable,yang2018graph,herzig2018mapping,gkanatsios2019attention}. 
Graph CNNs~\cite{kipf2016semi} have been used to propagate information between object and relationship nodes~\cite{yang2018graph,chen2019knowledge}. Some works also introduce generative models for scene understanding ~\cite{chen2019scene,gu2019scene}. 

\noindent \textbf{Contrastive Training:} Contrastive losses~\cite{arora2019theoretical} have been applied for scene understanding~\cite{Zhang2019:GCL}. They have also been applied for image captioning, visual question answering (VQA) and vector embeddings~\cite{yang2018shuffle,nagaraja2016modeling,rohrbach2016grounding}. They are based on the idea that it can be easier to randomly select similar and dissimilar samples. Such losses can then be used even if no label information is available.

\noindent \textbf{Losses:} In Visual Genome~\cite{krishna2017visual}, many semantically meaningful relations are not labeled. Furthermore, relations that can have multiple labels, for instance \texttt{on} and \texttt{sit}, are usually labeled only once. Networks for visual relationship detection which use cross-entropy as training loss may encounter difficulties during training, since almost identical pairs of objects can have either label, or even none at all. 
To overcome the problem of such contradicting label information, margin-based~\cite{krishnaswamy2019combining} and contrastive~\cite{Zhang2019:GCL} losses have been proposed.

\noindent \textbf{Support Inference:} Inferring physical support, i.e.~which structures for instance carry others, e.g. \texttt{floor} $\rightarrow$ \texttt{table}, was investigated for object segmentation~\cite{Silberman2012:SupportInference}, instance segmentation~\cite{zhuo2017indoor}, and also for scene graph inference~\cite{yang2017support}.

\noindent \textbf{Proposed model:} The proposed model uses neither iterations, except for LSTM-cells, nor message passing, nor a Graph CNN. As most recent works do, language priors are included. We further use a standard loss based on cross-entropy. Unlike most state-of-the-art algorithms, we propose to use a new module which has never before been used for semantic image analysis. The ODE module~\cite{Chen2018:NODE} can be interpreted as learning a deep residual model. In contrast to residual networks, the ODE-module \emph{continuously} models its solutions according to a pre-defined, problem-specific precision. 

This idea is motivated by works on gas, water, energy and transport networks which need to be modeled by systems of Partial Differential Equations (PDE). In~\cite{Fuegenschuh2006:CombinatorialModels}, it was proposed to simplify such systems to obtain systems of Ordinary Differential Equations (ODE), add one or multiple optimization objectives and then use a Mixed-Integer Linear Program to compute the solution. That implies that, given an (M)ILP, we can always find a system of ODEs that can generate the same solution among others. Using a neural ODE~\cite{Chen2018:NODE}, we can thus learn a function of the system of ODEs. Due to the end-to-end training, that system further encodes  near-optimal assignment functions per image.



\begin{figure*}[ht!]
\centering
\includegraphics[width=0.99\linewidth]{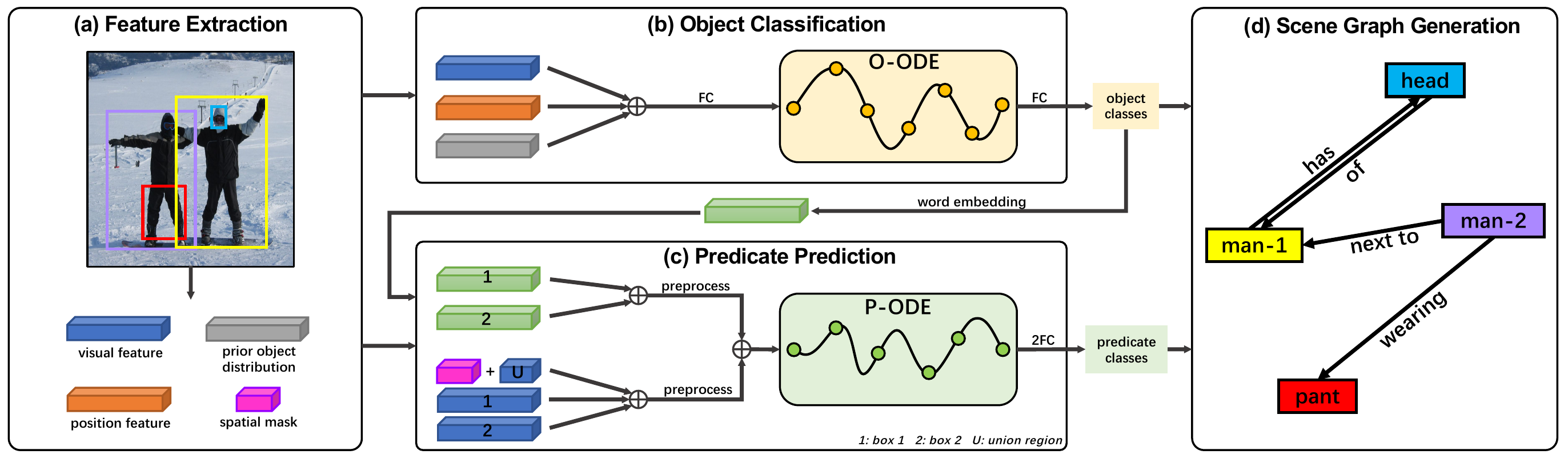}
\caption{Overview of the proposed method: 
(a) given an image, many object regions are proposed, (b) those regions are classified by an Object ODE (O-ODE), (c) after pre-processing the features of object pairs, a Predicate ODE (P-ODE) is applied to predict predicates, (d) a scene graph is generated.}
\label{fig:model}
\end{figure*}  

\section{Method} 
\label{Sec:method}

In this section, we first define scene graphs, then we define straight-forward ILP models for object and relationship classification in Sec.~\ref{Sec:VRD.ILP}. Neural Ordinary Differential Equations~\cite{Chen2018:NODE} are briefly explained in Sec.~\ref{Sec:NODE}. Both models are combined in Sec.~\ref{Sec:Assignments} and \ref{Sec:Archi}. 





\subsection{Scene Graphs}
A scene graph 
\cite{johnson2015image} is an abstract semantic representation of all objects in an image and the relationships between them. 
The set of objects 
consists of a set of bounding boxes 
along with a label for each object. 
The relationships between two objects, also known as predicates, 
consists of bounding boxes which usually cover both object bounding boxes, 
and labels for each relation. 
As in~\cite{zellers2018neural}, we include a particular predicate to indicate that there is no relation between two objects. 


\subsection{Models for Object and Relationship Detection}
\label{Sec:VRD.ILP}
Assume a graph $G_{obj}=(U_{obj},E_{obj})$ whose nodes represent the detected yet unlabeled objects in an image. Each node is further assigned to a label $l \in \cL_{obj}$. Each label $l$ is associated with a score $\alpha_{u,l}$ with $u \in U$. That score can represent an agreement with some given feature map. 
The integer variables $x_{u,l} \in \{0,1\}$ indicate that node $u$ is given label $l$. We assume that each object can only belong to a single class.

We further include a term $\beta_{u,u^\prime,l,l^\prime}$ which models statistical prior knowledge about co-occurrences of object pairs $u$ and $u^\prime$, $u \neq u^\prime$, and their corresponding labels $l$ and $l^\prime$. We then arrive at the following Integer Linear Problem (ILP) for object classification
\begin{subequations}
\begin{gather}
    \max \sum \limits_{u \in U_{obj}} \sum \limits_{l \in \cL_{obj}} \alpha_{u,l} x_{u,l} + w_{obj} \sum \limits_{u,\, u^\prime \in U_{obj}} \sum \limits_{l,\, l^\prime} \beta_{u,u^\prime,l,l^\prime} x_{u,l} x_{u^\prime,l^\prime} \\
    s.t. \quad x_{u,l} \in \{0,1\}, \quad \sum \limits_{l \in \cL_{obj}} x_{u,l} \leq 1
\end{gather}
\label{Eq:ILP.obj}
\end{subequations}
where $w_{obj}$ is a scalar weight which balances the two terms. How to exactly determine the scores $\alpha_{u,l}$, $\beta_{u,u^\prime,l,l^\prime}$ and the weight $w_{obj}$ constitutes a hyper-parameter search. The large search space makes this problem by itself challenging. Some of the parameters remain invariant for all images.

Assume a further graph $G_{pred}=(V_{pred},E_{pred})$ whose nodes represent all possible subject-object pairs. Each node is assigned to a label $k \in \cL_{pred}$. Each label $k$ is associated with a score $\alpha_{v,k}$ with $v \in V_{pred}$. 
The integer variables $x_{v,k} \in \{0,1\}$ indicate that node $v$ is given label $k$, for instance the subject-object pair \emph{dog-street} with a label \emph{sit} or \emph{walk}. The number of labels per node is limited by $T_v$, and the total number of labels by $K$. The latter arises from the \emph{recall-at-K} metric commonly used in visual relationship detection and scene graph estimation tasks. 

Here, we also include a term $\beta_{v,v^\prime,k,k^\prime}$ which models statistical prior knowledge about co-occurrences of subject-object pairs $v$ with pairs $v^\prime$, $v \neq v^\prime$, and their corresponding relationship labels $k$ and $k^\prime$. We then arrive at the following Integer Linear Problem (ILP) for relationship classification
\begin{subequations}
\begin{gather}
    \max \sum \limits_{v \in V_{pred}} \sum \limits_{k \in \cL_{pred}} \alpha_{v,k} x_{v,k} + w_{pred} \sum \limits_{v,\, v^\prime \in V_{pred}} \sum \limits_{k,\, k^\prime} \beta_{v,v^\prime,k,k^\prime} x_{v,k} x_{v^\prime,k^\prime} \\
    s.t. \quad x_{v,k} \in \{0,1\}, \quad \sum \limits_{k \in \cL} x_{v,k} \leq T_v, \quad 
    \sum \limits_{v \in V_{pred}} \sum \limits_{k \in \cL_{pred}} x_{v,k} \leq K,
\end{gather}
\label{Eq:ILP.pred}
\end{subequations}
where $w_{pred}$ is a scalar weight which balances the two terms. 

Denote by $\alpha(t)$ the \emph{function} in the single variable $t$ that assigns \emph{continuous} scalar weights which express label strengths. The values that the label weights take on vary with $t$.  
Likewise, let $\beta(t)$ be a function in $t$ that assigns scalar weights according to co-occurrence. From~\cite{Fuegenschuh2006:CombinatorialModels}, we can see that the optimization problems defined by Eqs.~\eqref{Eq:ILP.obj} and \eqref{Eq:ILP.pred} correspond to ordinary differential equation (ODE) models defined by
\begin{equation}
    \frac{d}{dt} f = f(x(t), \, t).
    \label{Eq:ODE}
\end{equation}
In other words, we can use Eqs.~\eqref{Eq:ILP.obj} and \eqref{Eq:ILP.pred} to obtain a solution of problem~\eqref{Eq:ODE} subject to several constraints. A disadvantage of the optimization problem in Eqs.~\eqref{Eq:ILP.obj} and \eqref{Eq:ILP.pred} is that it is not possible to backpropagate through the ILPs. Thus, feature-generating neural networks cannot be trained to adapt to given data when using the ILPs.


\subsection{Neural Ordinary Differential Equations~\cite{Chen2018:NODE}} 
\label{Sec:NODE} 
For many problems it is hard to explicitly define functions $\alpha(t)$ and $\beta(t)$. In \cite{Chen2018:NODE}, it was proposed to parameterize $f(t)$ by a function $f_\theta(t)$ based on neural network with parameters $\theta$. The idea is then to use $f_\theta(t)$ to solve an ODE. 
Thus, starting with an input $x(0)$ at time $t=0$, the output at time $T$ can be computed by a standard ODE-solver 
\begin{equation}
    \frac{d}{dt} f_\theta = f_\theta \left( \, x(t), \, t \right).
    \label{Eq:NODE}
\end{equation}
The dynamics of the system can thereby be approximated up to the required precision. Importantly, model states at times $0<t<T$ need not be evenly spaced. 



For back-propagation, the derivatives for the adjoint $a(t) = -\partial L / \partial x(t)$ 
\begin{equation}
    \frac{d}{d t} a = -a(t)^\top \frac{\partial}{\partial t} f ( \, x(t), \, t, \, \theta \, )
\end{equation}
and
\begin{equation}
    \frac{d}{d \theta} L = \int \limits_T^0 a(t)^\top \frac{\partial}{\partial \theta} f ( \, x(t), \, t, \, \theta \, ) \, dt
\end{equation}
have to be computed for loss $L$. This computation, along with the reverse computation of $x(t)$ starting at time $T$ can be done by a single call of the ODE-solver backwards in time. 


\subsection{Assignments by Neural Ordinary Differential Equations}
\label{Sec:Assignments}

While Eqs.~\eqref{Eq:ILP.obj} and \eqref{Eq:ILP.pred} can be used to obtain solutions of problem~\eqref{Eq:ODE}, this does not allow to train previous modules if used in a neural network. Furthermore, the hyper-parameters of the ILP must be determined in advance and cannot be easily adapted to the data. On the other hand, we can construct an ODE from Eqs.~\eqref{Eq:ILP.obj} or \eqref{Eq:ILP.pred}, respectively. Since a manual construction would be hard, we can use a neural ODE layer to learn the optimal assignment function. In contrast to an ILP, the neural ODE both allows to train a feature generating network in front of it, and can adapt to each image. The latter implies that the hyper-parameters of the corresponding, yet unknown ILP vary with each image.

ODEs involve the variable $t$ which is usually associated with time in many natural processes. Here, we are posed with the question what this variable represents for object or relationship classification. It will be demonstrated in Sec.~\ref{subsec:time} that the two variables of two neural ODE layers control how many objects and relationships are classified correctly (cf.~Fig.~\ref{fig:ode_vary}). In other words, specifying particular values of the variable of the relationship module  determines the connectivity of estimated scene graph, whereas such values of the object layer determine if too few or too many objects are correctly labeled.

As outlined in Fig.~\ref{fig:model}, we use two separate ODE layers. The visual features of detected objects, their positional features and the prior object distributions are concatenated into a vector $x_v$ and processed by an ODE layer in the object classifier
\begin{equation}
    \frac{d}{dt_1} f_{\theta_u} = f_{\theta_u} \left(  \, x_{u}(t_1), \, t_1 \right).
    \label{Eq:ODE.obj}
\end{equation}
In the following, this ODE layer will be denoted by \emph{O-ODE}.

The word embedding resulting from the object classifier, the spatial mask with the union box and the visual features of two detected and classified object are concatenated and pre-processed to yield vector $x_{v,v^\prime}$ before being processed by an ODE layer for predicate classification (\emph{P-ODE})
\begin{equation}
    \frac{d}{dt_2} f_{\theta_v} = f_{\theta_v} \left(  \, x_{v,v^\prime}(t_2), \, t_2 \right).
    \label{Eq:ODE.pred}
\end{equation}

The variables $t_1$ and $t_2$ in Eqs.~\eqref{Eq:ODE.obj} and \eqref{Eq:ODE.pred} control how many objects or relations are labeled. In other words, graphs constructed using different $t$ and $t^\prime$, either for Eq.~\eqref{Eq:ODE.obj} or \eqref{Eq:ODE.pred}, result in scene graphs with differently many objects or relations correctly labeled. 

\subsection{Architecture}
\label{Sec:Archi}
Our model is built on Faster-RCNN~\cite{ren2015faster} which provides proposal boxes, feature maps and primary object distributions. There are two fundamental modules in the model: object classifier and predicate classifier. Each of them contains a neural ordinary differential equation layer, Object ODE (O-ODE) and Predicate ODE (P-ODE). For both of them we use bidirectional LSTMs as the approximate function in the ODE solver. The data in the model are organized as sequences with random order.

In the object classifier, the feature maps, bounding box information and primary object distribution from Faster-RCNN are concatenated and fed through a fully connection layer. The object class scores are computed by the O-ODE and a following linear layer. The predicate classifier uses feature maps, spatial and semantic information of object pairs to predict the predicate categories. 
The spatial masks are forwarded into a group of convolution layers so that the output has the same size as the feature maps of the union box ($512 \cdot 7 \cdot 7$) and can be added per element. Global average pooling is applied on the feature maps of the subject box, object box and union box. The features of the subject, object, and union boxes are concatenated as  $(3 \cdot 512)$-dimensional visual vectors. Two $200$-dimensional semantic embedding vectors are generated from the subject and object classes predicted by the object classifier and concatenated as $400$-dimensional semantic vectors. 

The visual vectors and semantic vectors of object pairs can be pre-processed by three methods before the P-ODE:  
\emph{FC-Layer:} The $(3 \cdot 512)$-dimensional visual vectors and $400$-dimensional semantic vectors are forwarded into two independent fully connection layers that both have 512 neurons. Then, the outputs are concatenated together as $1024$-dimensional representation vectors for the P-ODE. 
\emph{GCNN:} The visual vectors and semantic vectors are first concatenated. Then, we use a graph convolutional neural network (GCNN) to infer information about context. Since the number of object pairs in each image is variable, we set each element on the diagonal of the adjacency matrix to $0.8$. The weight of $0.2$ is uniformly distributed among the remaining entries of each row. The output vectors of the GCNN are passed into the P-ODE. 
\emph{LSTM:} Similar as for the first variant, the $(3 \cdot 512)$-dimensional visual vectors and $400$-dimensional semantic vectors are fed into two single layer LSTMs. Both of them have the output dimension $512$. We concatenate the two outputs for the P-ODE. 

The final class scores of the relations are computed by the P-ODE followed by two linear layers.

\section{Experiments}
\label{Sec:Exps}

In this section, we firstly clarify the experimental settings
and implementation details. 
Then, we show quantitative and qualitative results on the Visual Genome (VG) benchmark dataset \cite{krishna2017visual} in terms of scene graph generation. 

\subsection{Dataset, Settings, and Evaluation}
\paragraph{Dataset}
We validated our methods on the VG benchmark dataset \cite{krishna2017visual} for the task of scene graph generation.
However, there are varying data pre-processing strategies and dataset splits in different works.
For fair comparison, we adopted the data split as described in~\cite{xu2017scene} which is the most widely used.
According to the data pre-processing strategy, the most-frequent 150 object categories and 50 predicate types are selected. The training set has $75651$ images while the test set has $32422$ images.

\paragraph{Settings}
For comparison with prior works, we use Faster R-CNN~\cite{ren2015faster} with VGG16~\cite{simonyan2014very} as the backbone network for proposing object candidates and extracting visual features.
We adopted the code base and a pre-trained model provided by~\cite{zellers2018neural}.
As in NeuralMotifs~\cite{zellers2018neural}, the input image is resized to $592\times592$, bounding box scales and dimension ratios are scaled, and a ROIAlign layer~\cite{He2017:MaskRCNN} is used to extract features within the boxes of object proposals and the union boxes of object pairs from the shared feature maps. 

\paragraph{Evaluation}
There are three standard experiment settings for evaluating the performance of scene graph generation:
(1) \textbf{Predicate classification} (PREDCLS): predict relationship labels of object pairs given ground truth bounding boxes and labels of objects.
(2) \textbf{Scene graph classification} (SGCLS):  given ground truth bounding boxes of objects, predict object labels and relationship labels. 
(3) \textbf{Scene graph detection} (SGGEN): predict boxes, labels of object proposals and relation labels of object pairs given an image.
Only when the labels of the subject, relation, and object are correctly classified, and the boxes of subject and object have more than $50\%$ intersection-over-union (IoU) with the ground truth, it is counted as a correctly detected entity.
The most widely adopted recall@K metrics ($K=[20,50,100]$) for relations are used to evaluate the system performance.

\paragraph{Training}
We train our model with the sum of the cross entropy losses for objects and predicates. We collect all annotated relationships in the image and add negative relationships so that the relation sequences in the batch have identical length  if there are sufficiently many ground truth boxes. We randomly select one if the object pairs are annotated with multiple relationships. For SGCLS and SGGEN, ground truth object labels are provided to the semantic part at the training stage. For fair comparison we use the same pre-trained Faster-RCNN as ~\cite{zellers2018neural} and freeze its parameters. An ADAM optimizer was used with batch size $6$, initial learning rate $10^{-4}$, and cross-entropy as loss both for object and relationship classification. 
We choose \emph{dopri5} as ODE solver and set the absolute tolerance $atol=0.01$ and the relative tolerance $rtol=0.01$. The integral time $t_{end}$ is set to $1.5$ (cf.~Sec.~\ref{subsec:time}). 
Please confer to the supplementary for a more detailed explanation of our model architecture.
\renewcommand{\arraystretch}{1} %
\begin{table*}[!ht]
\centering
\caption{Comparison on VG test set \cite{xu2017scene} \textbf{using graph constraints}. All numbers in \%.
We use the same object detection backbone provided by \cite{zellers2018neural} for fair comparison. Bold blue numbers indicate results better than competitors by $>0.5$. Regarding GCL~\cite{Zhang2019:GCL}, cf. to the text. 
Methods in the lower part use the same data split as \cite{xu2017scene}. 
Results for MSDN$\star$ \cite{li2017scene} are from \cite{zellers2018neural}.} 
\label{tab:results}
\begin{threeparttable}
\begin{tabular}{ccccccccccc}
\toprule
\multirowcell{2}{Data \\ Split}&\multirow{2}{*}{ Method } & \multicolumn{3}{c}{ SGGEN } & \multicolumn{3}{c}{ SGCLS } & \multicolumn{3}{c}{ PREDCLS } \cr
    \cmidrule(lr){3-5} \cmidrule(lr){6-8} \cmidrule(lr){9-11}
& &R@20 & R@50 & R@100 & R@20 & R@50 & R@100 & R@20 & R@50 & R@100\cr
\midrule
\multirow{2}{*}{\cite{li2017scene}}
&MSDN$\star$ \cite{li2017scene}	& - & 11.7 & 14.0 & - & 20.9 & 24.0 & - & 42.3 & 48.2 \\
&FacNet \cite{li2018factorizable}	& - & 13.1 & 16.5 & - & 22.8 & 28.6 & - & - & - \\
\midrule
\multirow{7}{*}{\rotatebox[origin=c]{90}{\cite{xu2017scene} split}}
&VRD \cite{lu2016visual}            &      &  0.3 &  0.5 &      & 11.8 & 14.1 &      & 27.9 & 35.0 \\
&IMP \cite{xu2017scene}             & 14.6 & 20.7 & 24.6 & 31.7 & 34.6 & 35.4 & 52.7 & 59.3 & 61.3 \\
&Graph R-CNN \cite{yang2018graph}   &    - & 11.4 & 13.7 &    - & 29.6 & 31.6 &    - & 54.2 & 59.1 \\
&Mem \cite{wang2019exploring}       &  7.7 & 11.4 & 13.9 & 23.3 & 27.8 & 29.5 & 42.1 & 53.2 & 57.9 \\
&MotifNet \cite{zellers2018neural}  & 21.4 & 27.2 & 30.3 & 32.9 & 35.8 & 36.5 & 58.5 & 65.2 & 67.1 \\
&MotifNet-Freq                      & 20.1 & 26.2 & 30.1 & 29.3 & 32.3 & 32.6 & 53.6 & 60.6 & 62.2 \\
&GCL~\cite{Zhang2019:GCL}           & 21.1          & 28.3 & 32.7 & 36.1 & 36.8 & 36.8 & 66.9 & 68.4 & 68.4 \\
&VCTREE \cite{Tang_2019_CVPR} & 22.0 & 27.9 & 31.3 & 35.2 & 38.1 & 38.8 & 60.1 & 66.4 & 68.1 \\
&CMAT \cite{chen2019counterfactual} & 22.1 & 27.9 & 31.2 & 35.9 & 39.0 & 39.8 & 60.2 & 66.4 & 68.1 \\

\midrule
&\textbf{ours-FC}                      & 21.5 & 27.5 & 30.9 & {\color{blue} \textbf{37.7} } & {\color{blue} \textbf{41.7} } & {\color{blue} \textbf{42.8} } & 58.6 & 66.1 & 68.1 \\
&\textbf{ours-GCNN}                     & 21.4 & 27.1 & 30.6 & 33.2 & 38.2 & 39.7 & 52.0 & 60.9 & 63.8 \\
&\textbf{ours-LSTM}                      & 21.6 & 27.7 & 31.0 & {\color{blue} \textbf{37.9}} & {\color{blue} \textbf{41.9}} & {\color{blue} \textbf{42.9} } & 58.9 & 66.0 & 67.9 \\
\bottomrule
\end{tabular}
\end{threeparttable}
\end{table*}

\subsection{Quantitative Results and Comparison}
\label{subsec:results}

Our results \textbf{using graph constraints} are shown in Tab.~\ref{tab:results}. The middle block indicates methods that all use the same data split that is used in~\cite{xu2017scene}. Two methods that use different splits are listed in the top section of the table. For MSD-net~\cite{li2017scene} in this part, we show the results reported in~\cite{zellers2018neural}. 

We used bold blue numbers to indicate the best result in any column that was at least $0.5\%$ larger than the next best competing method. 
Since GCL~\cite{Zhang2019:GCL} used a different VGG backbone than other works, and also larger input images than all other methods ($1024 \times 1024$ compared to $592 \times 592$), results cannot be fairly compared with other methods, so we did not highlight best results in this row (SGGEN-R@100 and PREDCLS-R@20).

The bottom part of Tab.~\ref{tab:results} shows results of the proposed ODE layers using three different front-ends: (1) One in which the function used inside the ODE layer is taken to be a linear layer (\textbf{ours-FC}), (2) the second in which a Graph-CNN is used (\textbf{ours-GCNN}), (3) and the third in which a  bidirectional LSTM is used (\textbf{ours-LSTM}).

As can be seen from Tab.~\ref{tab:results}, most recent methods (MotifNet~\cite{zellers2018neural}, VCTREE~\cite{Tang_2019_CVPR} and CMAT~\cite{chen2019counterfactual}) including ours are very similar in performance with respect to the scores in SGGEN and PREDCLS. 
The only exception among state-of-the-art works is GCL~\cite{Zhang2019:GCL} in two out of nine scores, yet by using a more powerful front-end. 

For SGCLS, however, the proposed ODE layer turns out to be very effective. Apparently both the version using a linear layer inside the ODE layer (\textbf{ours-FC}) and the one using the bidirectional LSTM (\textbf{ours-LSTM}) perform very similar. Their scores improve state-of-the-art (CMAT~\cite{chen2019counterfactual}) between $\approx 2\%$ (SGCLS-R@20), $\approx 3\%$ (SGCLS-R@50) and $\approx 3\%$ (SGCLS-R@100). Compared with GCL~\cite{Zhang2019:GCL} our results improve by up to $5\%$ (SGCLS-R@100) although our VGG was trained on coarser images. This demonstrates the effectiveness of the proposed ODE layer. It is sufficient to use a simple linear layer, since the ODE layer is so powerful that it produces similar outputs to those of a function based upon a more complicated and slower bidirectional LSTM.

\subsection{Ablation Studies}
\label{subsec:ablation}
For ablation studies, we consider several variants of the proposed network architecture. Results are shown in Tab.~\ref{tab:ablation}. For model-$1$ (first row), we removed the ODE networks and directly classified the output of the feature generating networks. This measures the impact of both ODE modules. In model-$2$ (second row), we removed only the ODE layer for object classification (O-ODE) in the proposed network. The third model shows the results when the relationship ODE (P-ODE) is removed. Finally, the fourth rows shows the results if both ODE modules are present. 
\renewcommand{\arraystretch}{1} 
\begin{table*}[!ht]
\centering
\caption{Ablation study demonstrating the effect if both ODE layers are removed (first row), only the layer for object classification (second row), only the layer for predicate classification (third row), or if both are present (last row).}
\label{tab:ablation}
\begin{threeparttable}
\begin{tabular}{c|cccccccc}
\toprule
\multirow{2}{*}{Model} & \multirow{2}{*}{O-ODE} &\multirow{2}{*}{P-ODE}  &\multicolumn{2}{c}{SGGEN} & \multicolumn{2}{c}{SGCLS} & \multicolumn{2}{c}{ PREDCLS } \cr 
 \cmidrule(lr){4-5} \cmidrule(lr){6-7} \cmidrule(lr){8-9}
& &								 & R@50 & R@100 & R@50 & R@100 & R@50 & R@100\cr
\midrule
1 & - 	    & -           & 27.1 &  30.4 &  35.3 &  36.1 &   65.4 &   67.4 \\
2 & - 	       & \checkmark	  &  27.3& 30.6 & 35.9 & 36.6 &  65.9 &  67.9 \\
3 & \checkmark & -            & 27.3 &  30.7&  41.4 &  42.5&  65.4 &   67.5 \\
4 & \checkmark & \checkmark   &   27.7 &  31.0 &  41.9 &  42.9 & 66.0 & 67.9 \\
\bottomrule
\end{tabular}
\end{threeparttable}
\end{table*}

As can be seen, removing the ODE layer for predicate classification has a negligible effect since the PREDCLS scores hardly change. Removing the ODE layer for object classification has a strong, negative effect, however. This study shows that the ODE module can have a very positive impact. It further confirms our main claim that classification/assignment problems can be solved by means of neural ODEs. 

Regarding the scores on PREDCLS, we conjecture that the noise in VG (missing relationship labels, incorrect labels) is so strong that scores cannot improve anymore. This hypothesis is supported by the fact that results in the past two years have not improved since~\cite{zellers2018neural}.

\subsection{Neural ODE Analysis}
\label{subsec:time}

We propose two neural ordinary differential equations: Object ODE (O-ODE) and Predicate ODE (P-ODE) in the object classifier and predicate classifier respectively. Here, we tune the hyper-parameter, integral time $t_{end}$ in the ODE solver, to understand how that variables  influences neural ordinary differential equations and final performance. Moreover, we extract the hidden states at different points in time and generate the scene graph to visualize how features are refined in the ODE space. 

\begin{figure*}[!ht]
    \centering
    \includegraphics[width=1.0\linewidth,height=4cm]{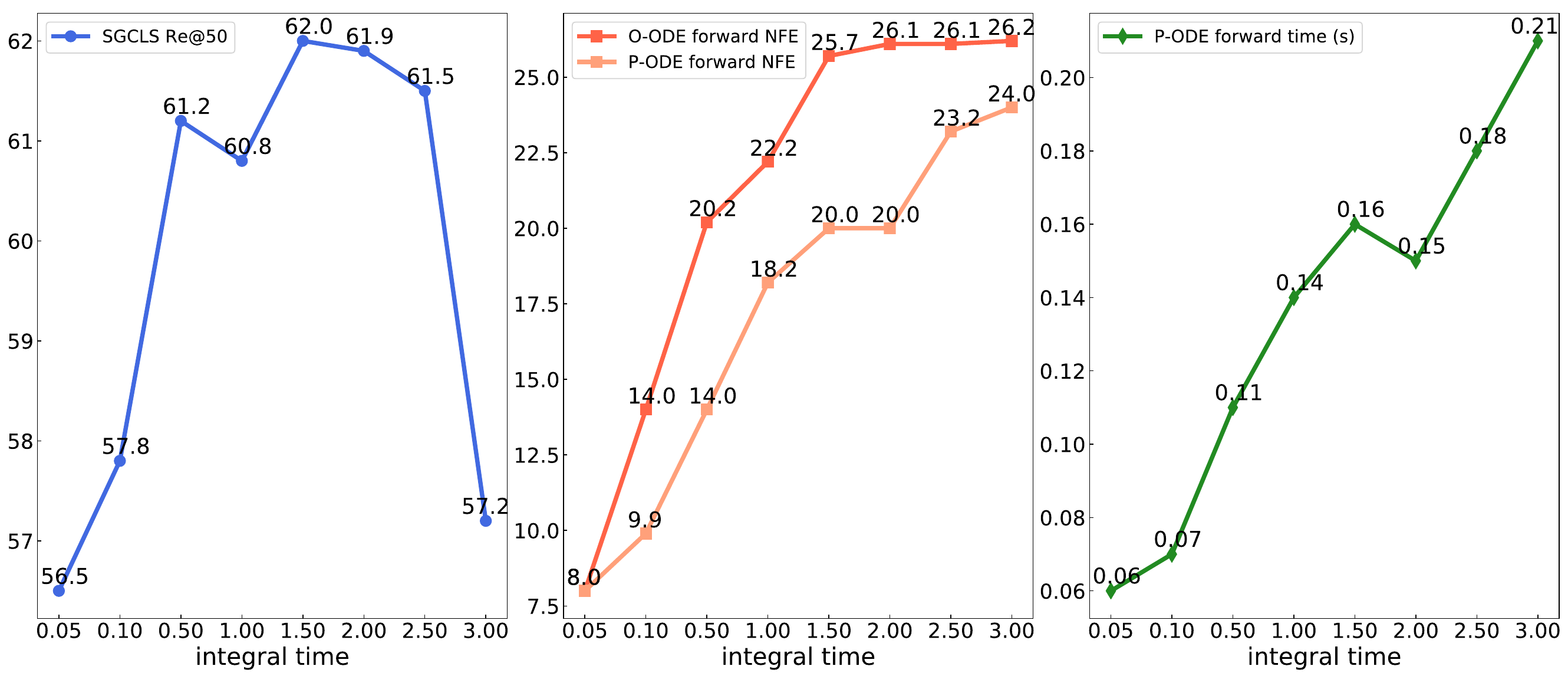}
    \caption{The effect of integral time in the ODE solver on SGCLS Re@50, average forward NFE (number of function evaluation) and average forward running time per image of O-ODE and P-ODE. Because of different number of objects in different images the experimental results are based on 5000 samples from the training set.} 
    \label{fig:ode_info}
\end{figure*}

Since the O-ODE and P-ODE both work in the setting SGCLS, SGCLS Re@50 is used as performance indicator. Because different images contain different number of objects, we randomly sample 5000 images from the training set to compute the average number of function evaluation (NFE) for a forward pass and the computation time per image. 
This experiment is implemented on a single GTX 1080Ti GPU. 
We vary the integral time $t_{end}$ from $0.05$ to $3.00$ (x-axes). 

According to the left plot in Fig.~\ref{fig:ode_info}, SGCLS Re@50 increases gradually from $56.5$ to $62.0$ until $T=1.50$ and then decreases; (middle plot) the average forward NFE of O-ODE increases from $8.0$ to $26.2$ and from $8.0$ to $24.0$ for P-ODE; (right plot) the average forward time of the P-ODE layer increases from $0.06$ seconds to $0.21$ seconds whereas the average forward time of the O-ODE layer is negligibly small ($<10^{-4}$ seconds), thus we omit these measurements. The P-ODE requires much more time than the O-ODE due to the large amount of object pairs. Considering the above points, we set $t_{end}=1.5$ for all other experiments.

\begin{figure*}[!ht]
    \centering
    \includegraphics[width=1.0\linewidth]{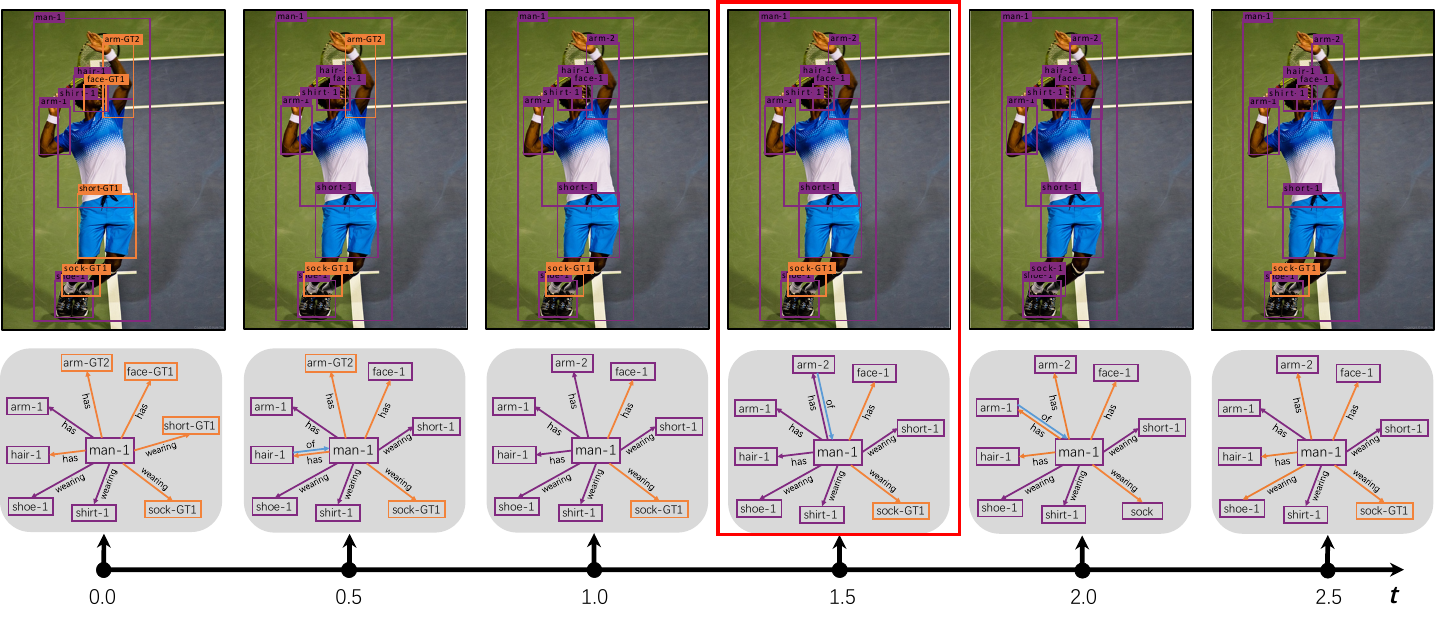}
    \caption{We evaluate the model trained with the hyper-parameter $t_{end}=1.5$, i.e.~during training the hidden vector at $t=1.5$ is used for classifications. The upper row visualizes detection results while the lower row shows the corresponding scene graphs. The red box indicates the state corresponding to $t=1.5$. \newline
    The orange boxes and edges indicate the objects and predicates in the ground truth that are not detected, purple that the objects and predicates are predicted correctly. Blue edges show false positives.}
    \label{fig:ode_vary}
\end{figure*}

To provide an intuitive interpretation of the parameters $t_1$ and $t_2$ in the O-ODE and the P-ODE, respectively, we use a model trained on SGGEN with the hyper-parameter $t_{end}=1.5$ and evaluate it using different latent vectors corresponding to different points between $0$ and $2.5$. 
The hidden states at $t=2.0$ and $t=2.5$ imply extrapolation. The detection results and scene graphs generated by the different hidden states are shown in Fig.~\ref{fig:ode_vary}. The features are more and more refined by the neural ODE layers if $t_1$ and $t_2$ increase, hence more objects and relationships are correctly classified. After $t_{end}=1.5$, 
increasingly many relationships are misclassified due to the extrapolation.

\subsection{Qualitative Results}
\label{subsec:qualitative}

\begin{figure*}[!ht]
    \centering
    \includegraphics[width=1.0\linewidth]{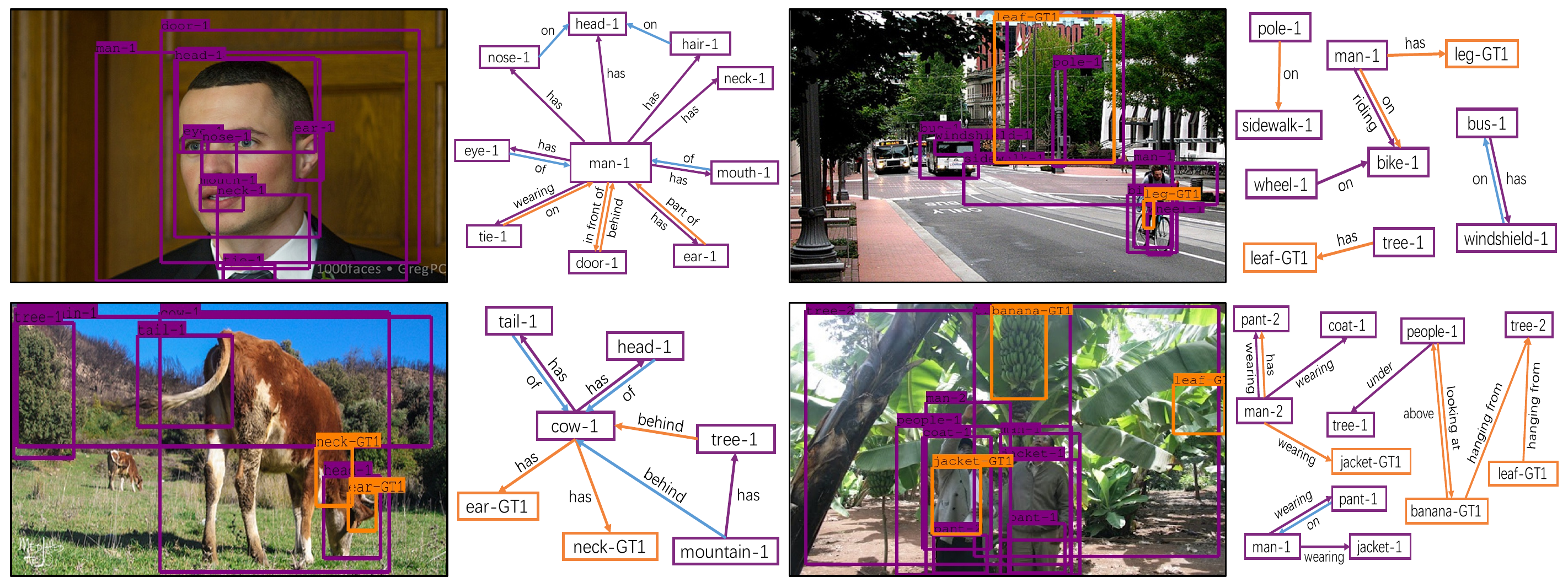}
    \caption{Qualitative results from our model in the scene graph generation setting. 
    Purple boxes denote correctly detected objects while orange boxes denote ground truth objects that are not detected.
    Purple edges correspond to correctly classified relationships at the R@20 setting while orange edges denote ground truth relationships that are not detected. Blue edges denote detected relationships that do not exist in ground truth annotations (false positives).
    }
    \label{fig:qualitative}
\end{figure*}

Qualitative results for scene graph generation (SGGEN) are shown in Fig.~\ref{fig:qualitative}. The images include object detections. The purple color indicates correctly detected and classified objects and relations, whereas orange means failure. The blue color indicates false negatives, i.e.~relationships that are not in the ground truth. 

Most of the errors stem from the object detection stage. Whenever an object is not detected, relationships connecting this object are also not present in the scene graph. The cause of these errors is the Faster R-CNN detector which is used in all previous works. 

There are a few false positives (blue links) which are semantically meaningful, for instance \texttt{eye}-\texttt{of}-\texttt{man} in the upper left example. In other words, the ground truth lacks this particular relationship in such cases. Several false positives result from semantically indistinguishable classes, for instance \texttt{has} and \texttt{of} in the lower left example. 

\section{Conclusions} 
\label{Sec:Concs}


We presented Neural Ordinary Differential Equations for Scene Understanding (NODIS). The idea of this work is based on the fact that Mixed-Integer Linear Programs can be used to solve problems defined by ordinary differential equations; therefore, given a particular (M)ILP, we can find a system of ODEs that can produce the solution of the (M)ILP within a time series. Since it is not possible to manually define the system of ODE to solve, we draw on recent advances in machine learning and use a trainable function approximator instead of an explicitly defined system of ODEs. 
In other words, the proposed network \emph{learns} the optimal function to solve the assignment problem, whereas previous works manually define modules to do so.

We use this newly defined module for object classification and relationship classification. 
The proposed model using ODE layers shows large improvements on SGCLS, between $2\%$ and $3\%$ compared with the best SOTA in that category. We believe that ODE layers can be valuable improvements for neural architectures in semantic image understanding. 
\section*{Acknowledgement}
This work was partially supported by the DFG grant COVMAP (RO 2497/12-2) and EXC 2122.

\clearpage
%
%
\bibliographystyle{splncs04}
\bibliography{ref}

\begin{thebibliography}{10}
\providecommand{\url}[1]{\texttt{#1}}
\providecommand{\urlprefix}{URL }
\providecommand{\doi}[1]{https://doi.org/#1}

\bibitem{arora2019theoretical}
Arora, S., Khandeparkar, H., Khodak, M., Plevrakis, O., Saunshi, N.: A
  theoretical analysis of contrastive unsupervised representation learning. In:
  Proceedings of Machine Learning Research (PMLR) (2019)

\bibitem{chen2019counterfactual}
Chen, L., Zhang, H., Xiao, J., He, X., Pu, S., Chang, S.F.: Counterfactual
  critic multi-agent training for scene graph generation. In: IEEE
  International Conference on Computer Vision (ICCV). pp. 4613--4623 (2019)

\bibitem{Chen2018:NODE}
Chen, T.Q., Rubanova, Y., Bettencourt, J., Duvenaud, D.: Neural {O}rdinary
  {D}ifferential {E}quations. In: Neural Information Processing Systems
  (NeurIPS). pp. 6571--6583 (2018)

\bibitem{chen2019knowledge}
Chen, T., Yu, W., Chen, R., Lin, L.: Knowledge-embedded routing network for
  scene graph generation. In: Proceedings of the IEEE Conference on Computer
  Vision and Pattern Recognition. pp. 6163--6171 (2019)

\bibitem{chen2019scene}
Chen, V.S., Varma, P., Krishna, R., Bernstein, M., Re, C., Fei-Fei, L.: Scene
  graph prediction with limited labels. In: Proceedings of the IEEE
  International Conference on Computer Vision. pp. 2580--2590 (2019)

\bibitem{dai2017detecting}
Dai, B., Zhang, Y., Lin, D.: Detecting visual relationships with deep
  relational networks. In: IEEE Conference on Computer Vision and Pattern
  Recognition (CVPR). pp. 3076--3086 (2017)

\bibitem{divvala2009empirical}
Divvala, S.K., Hoiem, D., Hays, J.H., Efros, A.A., Hebert, M.: An empirical
  study of context in object detection. In: IEEE Conference on Computer Vision
  and Pattern Recognition (CVPR). pp. 1271--1278 (2009)

\bibitem{Fuegenschuh2006:CombinatorialModels}
Fügenschuh, A., Herty, M., Klar, A., Martin, A.: Combinatorial and
  {C}ontinuous {C}odels for the {O}ptimization of {T}raffic {F}lows on
  {N}etworks. SIAM Journal on Optimization  \textbf{16}(4),  1155--1176 (2006)

\bibitem{gkanatsios2019attention}
Gkanatsios, N., Pitsikalis, V., Koutras, P., Maragos, P.:
  Attention-translation-relation network for scalable scene graph generation.
  In: Proceedings of the IEEE International Conference on Computer Vision
  Workshops. pp.~0--0 (2019)

\bibitem{gu2019scene}
Gu, J., Zhao, H., Lin, Z., Li, S., Cai, J., Ling, M.: Scene graph generation
  with external knowledge and image reconstruction. In: Proceedings of the IEEE
  Conference on Computer Vision and Pattern Recognition. pp. 1969--1978 (2019)

\bibitem{He2017:MaskRCNN}
He, K., Georgia, G., Doll\'ar, P., Girshick, R.: Mask {R-CNN}. In: IEEE
  International Conference on Computer Vision (ICCV). pp. 2961--2969 (2017)

\bibitem{herzig2018mapping}
Herzig, R., Raboh, M., Chechik, G., Berant, J., Globerson, A.: Mapping images
  to scene graphs with permutation-invariant structured prediction. In:
  Advances in Neural Information Processing Systems. pp. 7211--7221 (2018)

\bibitem{hu2018relation}
Hu, H., Gu, J., Zhang, Z., Dai, J., Wei, Y.: Relation networks for object
  detection. In: IEEE Conference on Computer Vision and Pattern Recognition
  (CVPR). pp. 3588--3597 (2018)

\bibitem{hu2019exploiting}
Hu, T., Liao, W., Yang, M.Y., Rosenhahn, B.: Exploiting attention for visual
  relationship detection. In: German Conference on Pattern Recognition. pp.
  331--344. Springer (2019)

\bibitem{johnson2015image}
Johnson, J., Krishna, R., Stark, M., Li, L.J., Shamma, D., Bernstein, M.,
  Fei-Fei, L.: Image retrieval using scene graphs. In: IEEE Conference on
  Computer Vision and Pattern Recognition (CVPR). pp. 3668--3678 (2015)

\bibitem{kipf2016semi}
Kipf, T.N., Welling, M.: Semi-supervised classification with graph
  convolutional networks. In: International Conference on Learning
  Representations (ICLR) (2016)

\bibitem{kluger2019temporally}
Kluger, F., Ackermann, H., Yang, M.Y., Rosenhahn, B.: Temporally consistent
  horizon lines. In: ICRA (2020)

\bibitem{kluger2020consac}
Kluger, F., Brachmann, E., Ackermann, H., Rother, C., Yang, M.Y., Rosenhahn,
  B.: Consac: Robust multi-model fitting by conditional sample consensus. In:
  IEEE/CVF Conference on Computer Vision and Pattern Recognition. pp.
  4634--4643 (2020)

\bibitem{krishna2018referring}
Krishna, R., Chami, I., Bernstein, M., Fei-Fei, L.: Referring relationships.
  In: IEEE Conference on Computer Vision and Pattern Recognition (CVPR) (2018)

\bibitem{krishna2017visual}
Krishna, R., Zhu, Y., Groth, O., Johnson, J., Hata, K., Kravitz, J., Chen, S.,
  Kalantidis, Y., Li, L.J., Shamma, D.A., et~al.: Visual genome: Connecting
  language and vision using crowdsourced dense image annotations. International
  Journal on Computer Vision (IJCV)  \textbf{123}(1),  32--73 (2017)

\bibitem{krishnaswamy2019combining}
Krishnaswamy, N., Friedman, S., Pustejovsky, J.: Combining deep learning and
  qualitative spatial reasoning to learn complex structures from sparse
  examples with noise. In: Association for the Advancement of Artificial
  Intelligence (AAAI). vol.~33, pp. 2911--2918 (2019)

\bibitem{ladicky2010graph}
Ladicky, L., Russell, C., Kohli, P., Torr, P.H.: Graph cut based inference with
  co-occurrence statistics. In: European Conference on Computer Vision (ECCV).
  pp. 239--253. Springer (2010)

\bibitem{li2017vip}
Li, Y., Ouyang, W., Wang, X.: Vip-cnn: Visual phrase guided convolutional
  neural network. In: IEEE Conference on Computer Vision and Pattern
  Recognition (CVPR). pp. 1347--1356 (2017)

\bibitem{li2018factorizable}
Li, Y., Ouyang, W., Zhou, B., Shi, J., Zhang, C., Wang, X.: Factorizable net:
  an efficient subgraph-based framework for scene graph generation. In:
  European Conference on Computer Vision (ECCV). pp. 346--363. Springer (2018)

\bibitem{li2017scene}
Li, Y., Ouyang, W., Zhou, B., Wang, K., Wang, X.: Scene graph generation from
  objects, phrases and region captions. In: IEEE International Conference on
  Computer Vision (ICCV). pp. 1261--1270 (2017)

\bibitem{liang2017deep}
Liang, X., Lee, L., Xing, E.P.: Deep variation-structured reinforcement
  learning for visual relationship and attribute detection. In: IEEE
  International Conference on Computer Vision (ICCV). pp. 848--857 (2017)

\bibitem{Liao_2019_CVPR_Workshops}
Liao, W., Rosenhahn, B., Shuai, L., Yang, M.Y.: Natural language guided visual
  relationship detection. In: IEEE/CVF Conference on Computer Vision and
  Pattern Recognition (CVPR) Workshops (2019)

\bibitem{liu2018structure}
Liu, Y., Wang, R., Shan, S., Chen, X.: Structure inference net: Object
  detection using scene-level context and instance-level relationships. In:
  IEEE Conference on Computer Vision and Pattern Recognition (CVPR). pp.
  6985--6994 (2018)

\bibitem{lu2016visual}
Lu, C., Krishna, R., Bernstein, M., Fei-Fei, L.: Visual relationship detection
  with language priors. In: European Conference on Computer Vision (ECCV). pp.
  852--869 (2016)

\bibitem{mikolov2013efficient}
Mikolov, T., Chen, K., Corrado, G., Dean, J.: Efficient estimation of word
  representations in vector space. arXiv:1301.3781  (2013)

\bibitem{nagaraja2016modeling}
Nagaraja, V., Morariu, V., Davis, L.: Modeling context between objects for
  referring expression understanding. In: European Conference on Computer
  Vision (ECCV). pp. 792--807 (2016)

\bibitem{reinders2019learning}
Reinders, C., Ackermann, H., Yang, M.Y., Rosenhahn, B.: Learning convolutional
  neural networks for object detection with very little training data. In:
  Multimodal Scene Understanding, pp. 65--100. Elsevier (2019)

\bibitem{ren2015faster}
Ren, S., He, K., Girshick, R., Sun, J.: Faster {R-CNN}: Towards real-time
  object detection with region proposal networks. In: Neural Information
  Processing Systems (NeurIPS). pp. 91--99 (2015)

\bibitem{rohrbach2016grounding}
Rohrbach, A., Rohrbach, M., Hu, R., Darrell, T., Schiele, B.: Grounding of
  textual phrases in images by reconstruction. In: European Conference on
  Computer Vision (ECCV). pp. 817--834. Springer (2016)

\bibitem{shi2019explainable}
Shi, J., Zhang, H., Li, J.: Explainable and explicit visual reasoning over
  scene graphs. In: IEEE Conference on Computer Vision and Pattern Recognition
  (CVPR). pp. 8376--8384 (2019)

\bibitem{Silberman2012:SupportInference}
Silberman, N., Hoiem, D., Kohli, P., Fergus, R.: Indoor {S}egmentation and
  {S}upport {I}nference from {RGBD} {I}mages. In: European Conference on
  Computer Vision (ECCV). pp. 746--760 (2012)

\bibitem{simonyan2014very}
Simonyan, K., Zisserman, A.: Very deep convolutional networks for large-scale
  image recognition. arXiv:1409.1556  (2014)

\bibitem{Tang_2019_CVPR}
Tang, K., Zhang, H., Wu, B., Luo, W., Liu, W.: Learning to compose dynamic tree
  structures for visual contexts. In: IEEE Conference on Computer Vision and
  Pattern Recognition (CVPR) (June 2019)

\bibitem{teney2017graph}
Teney, D., Liu, L., van~den Hengel, A.: Graph-structured representations for
  visual question answering. In: IEEE Conference on Computer Vision and Pattern
  Recognition (CVPR). pp. 3233--3241 (2017)

\bibitem{wang2019exploring}
Wang, W., Wang, R., Shan, S., Chen, X.: Exploring context and visual pattern of
  relationship for scene graph generation. In: IEEE Conference on Computer
  Vision and Pattern Recognition (CVPR). pp. 8188--8197 (2019)

\bibitem{xu2017scene}
Xu, D., Zhu, Y., Choy, C.B., Fei-Fei, L.: Scene graph generation by iterative
  message passing. In: IEEE Conference on Computer Vision and Pattern
  Recognition (CVPR). pp. 5410--5419 (2017)

\bibitem{yang2018graph}
Yang, J., Lu, J., Lee, S., Batra, D., Parikh, D.: Graph r-cnn for scene graph
  generation. In: European Conference on Computer Vision (ECCV). pp. 690--706
  (2018)

\bibitem{yang2017support}
Yang, M.Y., Liao, W., Ackermann, H., Rosenhahn, B.: On support relations and
  semantic scene graphs. ISPRS Journal of Photogrammetry and Remote Sensing
  (ISPRS)  \textbf{131},  15--25 (2017)

\bibitem{yang2019auto}
Yang, X., Tang, K., Zhang, H., Cai, J.: Auto-encoding scene graphs for image
  captioning. In: IEEE Conference on Computer Vision and Pattern Recognition
  (CVPR). pp. 10685--10694 (2019)

\bibitem{yang2018shuffle}
Yang, X., Zhang, H., Cai, J.: Shuffle-then-assemble: Learning object-agnostic
  visual relationship features. In: European Conference on Computer Vision
  (ECCV). pp. 36--52 (2018)

\bibitem{yao2010modeling}
Yao, B., Fei-Fei, L.: Modeling mutual context of object and human pose in
  human-object interaction activities. In: IEEE Conference on Computer Vision
  and Pattern Recognition (CVPR). pp. 17--24 (2010)

\bibitem{yao2018exploring}
Yao, T., Pan, Y., Li, Y., Mei, T.: Exploring visual relationship for image
  captioning. In: European Conference on Computer Vision (ECCV), pp. 711--727.
  Springer (2018)

\bibitem{yu2017visual}
Yu, R., Li, A., Morariu, V.I., Davis, L.S.: Visual relationship detection with
  internal and external linguistic knowledge distillation. In: IEEE
  International Conference on Computer Vision (ICCV). pp. 1974--1982 (2017)

\bibitem{zellers2018neural}
Zellers, R., Yatskar, M., Thomson, S., Choi, Y.: Neural motifs: Scene graph
  parsing with global context. In: IEEE Conference on Computer Vision and
  Pattern Recognition (CVPR). pp. 5831--5840 (2018)

\bibitem{Zhang2019:GCL}
Zhang, J., Shih, K.J., Elgammal, A., Tao, A., Catanzaro, B.: Graphical
  contrastive losses for scene graph parsing. In: IEEE Conference on Computer
  Vision and Pattern Recognition (CVPR) (2019)

\bibitem{zhuang2017towards}
Zhuang, B., Liu, L., Shen, C., Reid, I.: Towards context-aware interaction
  recognition for visual relationship detection. In: IEEE International
  Conference on Computer Vision (ICCV). pp. 589--598 (2017)

\bibitem{zhuo2017indoor}
Zhuo, W., Salzmann, M., He, X., Liu, M.: Indoor scene parsing with instance
  segmentation, semantic labeling and support relationship inference. In: IEEE
  Conference on Computer Vision and Pattern Recognition (CVPR). pp. 5429--5437
  (2017)

\end{thebibliography}
\end{document}